\documentclass[10pt,twocolumn]{article}

\usepackage[letterpaper,margin=0.68in,columnsep=0.24in]{geometry}
\usepackage[utf8]{inputenc}
\pdfmapfile{+cm.map}
\pdfmapfile{+cmextra.map}
\pdfmapfile{+symbols.map}
\pdfmapline{+tcrm1000 CMR10 <cmr10.pfb}
\usepackage{amsmath,amssymb}
\usepackage{booktabs,tabularx,array}
\usepackage{enumitem}
\usepackage{listings}
\usepackage{xcolor}
\usepackage{url}
\usepackage[hidelinks]{hyperref}

\setlist[itemize]{leftmargin=*,nosep}
\setlist[enumerate]{leftmargin=*,nosep}
\newcommand{\ContextPipe}{\textsc{ContextPipe}}
\newcommand{\ContextSources}{\textsc{ContextSources}}
\newcommand{\PipelineStats}{\textsc{PipelineStats}}
\newcommand{\EmergentContext}{\textsc{EmergentContext}}
\newcommand{\SessionLatches}{\textsc{SessionLatches}}
\newcommand{\OptimizeLimits}{\textsc{OptimizeLimits}}
\newcommand{\RecoveryState}{\textsc{RecoveryState}}
\newcommand{\ProviderCachePolicy}{\textsc{ProviderCachePolicy}}
\newcommand{\code}[1]{\texttt{#1}}
\newcommand{\heading}[1]{\medskip\noindent\textit{#1}}
\title{\bfseries ContextPipe: Database-Inspired Context Assembly for\\Long-Horizon Agents}
\author{Peng Xu \and Zuyu Zhang \and Yuze Sun \and Feng Tian \and Long Wang \and Chen Zhang\\
\small MatrixOrigin \quad Tsinghua University\\
\small \texttt{\{xupeng; zuyuzhang; sunyuze\}@matrixorigin.cn}\\
\small \texttt{\{tianfeng; wanglong\}@matrixorigin.io} \quad \texttt{fortunelink\_chris@163.com}}
\date{}

\begin{document}
\maketitle

\begin{abstract}
Long-horizon large language model (LLM) agents require context assembly: the runtime must decide what to include in each prompt, in what order, and when to compact history under a hard context-window budget and a byte-sensitive prompt cache. In production agentic systems, this logic is scattered across prompt builders, ad hoc compaction routines, cache-break workarounds, and per-provider shims. We argue that context assembly is structurally isomorphic to query execution in a relational database: both execute under a hard budget, exploit a tiered cache, and leverage statistics. We adopt this discipline in \ContextPipe: a five-phase pipeline (Plan $\rightarrow$ Bind $\rightarrow$ Optimize $\rightarrow$ Execute $\rightarrow$ Feedback) backed by a structured data-source catalog, a deterministic cache-aware optimizer, and an EXPLAIN ANALYZE trace. We show the context in \ContextPipe{} is auditable, replayable, and failure-isolated. A preliminary evaluation using the SWE-bench Pro Qutebrowser subset shows that, compared to the append-only context construction policy, \ContextPipe{} reduces total token volume by 31\%, LLM calls by 23\%, and response time by 9\% at a cost in lower KV cache-hit ratio.
\end{abstract}

\noindent\textbf{VLDB Workshop Reference Format:} Peng Xu, Zuyu Zhang, Yuze Sun, Feng Tian, Long Wang, Chen Zhang. ContextPipe: Database-Inspired Context Assembly for Long-Horizon Agents. VLDB 2026 Workshop: ADS 2026: The Joint Workshop on Agentic Data Systems and Data-Centric AI (The 1st ADS \& 3rd DATAI).

\section{Introduction}

A large language model (LLM) agent runs in a loop: it receives a user message, assembles a prompt, calls the model, interprets the response (often a tool call), executes any tools, and feeds the result into the next turn [21, 27]. Each turn is fundamentally a request-assembly problem: the runtime must collect system prompts, tool schemas, conversation history, retrieved memory, skill instructions, and runtime identity into one API call whose input size cannot exceed the model's context window, under latency and cost pressure from providers that charge per token and cache byte-identical prefixes [2, 19]. Raw concatenation works for short exchanges, but it fails once histories or tool results dominate the prompt; beyond that point the runtime must actively decide what remains, what is summarized, and what is deferred [20], and it must do so again on every turn of a long-horizon task.

In practice, production agentic systems have accumulated a large set of context-handling components to make these decisions: prompt builders per session, compaction routines for history, provider-specific cache markers, and attachment injection for tool output. The result resembles a pre-optimizer database inside the agent runtime: each site makes local decisions that are individually reasonable but collectively opaque, and two of those decisions carry outsized risk. First, caching imposes a byte-stability constraint: automatic prefix caching [19, 29] reuses the longest byte-identical prefix shared with a prior request, so appending after a cached prefix is inexpensive but a one-byte change anywhere in a stable block invalidates the suffix and forces full recomputation of the prefix [14]. Second, ordering is consequential even where caching is unaffected: long-context models do not attend uniformly across the prompt, and content placed in the middle is systematically under-attended relative to content near the beginning or end [16], so section order affects output quality as well as cost---putting volatile memory late can protect the cache while also changing whether the model uses that memory in the current turn. Typical runtimes cannot explain why the assembler chose a particular context, what a better choice would have cost, or whether a cache broke because of a tool-schema edit, a session-header change, or compacted placeholder drift; there is no catalog of context sources by lifecycle, no deterministic policy that jointly handles cache placement and compaction, and no trace explaining what happened and why.

We observe that this resembles the problem that relational database systems solved half a century ago [10, 22]. A database optimizer takes a declarative intent (a SQL query) and a catalog (tables, indexes, statistics), produces a physical plan that respects hard resource limits (memory and I/O), exploits a tiered cache (buffer pool), and exposes its reasoning (EXPLAIN ANALYZE). A context assembler has a declarative intent (the user input message), an implicit catalog (session states, available sources and their lifecycles and costs), a hard resource limit (the context window), and a tiered cache (the prompt prefix cache). However, it typically lacks a formal structure that makes these analogues visible.

This paper presents \ContextPipe, a runtime that plans, optimizes, and executes LLM context construction using database-inspired execution principles. Its contributions are:

\begin{itemize}
\item A pipeline abstraction for LLM context assembly with five explicit phases: Plan, Bind, Optimize, Execute, and Feedback. These phases mirror the shape of a query executor and unify functionality that is often distributed across many scattered modules (Section 2).
\item A deterministic, cache-aware optimizer that frames context assembly as an ordered, budget-constrained placement problem and solves it using cache-aware ordered placement: a fixed ascending-volatility sort and tier-gated, individually-gated compaction transforms, every one of which is recorded for audit whether applied or skipped (Section 3).
\item A structured data-source catalog (\ContextSources) that organizes the session states in a typical agent loop into eight lifecycle tiers, two of which are \SessionLatches{} and \EmergentContext, to make otherwise implicit behaviors first-class (Section 4).
\item Two mechanisms that let the pipeline adapt across turns instead of reacting only to the current prompt: predictive context pressure in Plan, which sizes the next response from historical percentiles rather than a fixed reserve, and TTL-guarded emergent context across Bind and Feedback, which safely carries mid-turn discoveries (e.g., a tool result that triggers a new skill) into the next turn without staleness or duplication (Sections 2, 4).
\end{itemize}

The rest of this paper is organized as follows. Section 2 presents the architecture. Section 3 presents the optimizer, Section 4 describes the catalog design, Section 5 presents the trace format, and Section 6 examines the implementation techniques. Section 7 reports the empirical evaluation between \ContextPipe{} and append-only context construction on long-horizon agent tasks. Section 8 covers previous work on context assembly, and Section 9 concludes.

\section{The ContextPipe Runtime}

This section describes the \ContextPipe{} runtime and its five-phase execution model. \ContextPipe{} treats one LLM turn as a single invocation of a staged pipeline:

\[
\text{Plan} \rightarrow \text{Bind} \rightarrow \text{Optimize} \rightarrow \text{Execute} \rightarrow \text{Feedback}.
\]

Each phase has a well-defined input type, output type, and purity constraint, and the phases together are the only path by which an LLM call is produced. There are no parallel payload builders for subruns, skill runs, compaction agents, or forked children; all of these invocations run through the same pipeline with different \ContextSources{} inputs.

\subsection{Overview}

\begin{figure}[t]
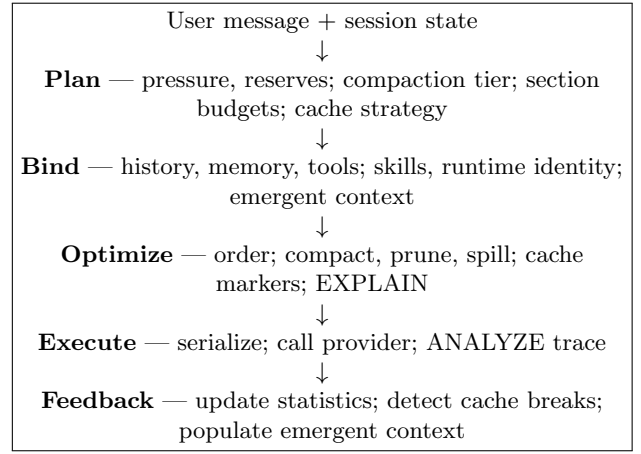

\centering
\fbox{\begin{minipage}{0.91\columnwidth}\centering\small
User message + session state\\$\downarrow$\\
\textbf{Plan} --- pressure, reserves; compaction tier; section budgets; cache strategy\\$\downarrow$\\
\textbf{Bind} --- history, memory, tools; skills, runtime identity; emergent context\\$\downarrow$\\
\textbf{Optimize} --- order; compact, prune, spill; cache markers; EXPLAIN\\$\downarrow$\\
\textbf{Execute} --- serialize; call provider; ANALYZE trace\\$\downarrow$\\
\textbf{Feedback} --- update statistics; detect cache breaks; populate emergent context
\end{minipage}}
\caption{\ContextPipe{} routes every LLM call through one auditable five-phase assembly path. Plan is pure; Bind is the only phase that performs external I/O; Execute is the only phase that contacts the model. Feedback populates emergent context for the next turn.}
\label{fig:pipeline}
\end{figure}

Figure~\ref{fig:pipeline} sketches the data flow. Plan is pure computation over the catalog and accumulated statistics: it decides what the turn should contain, how much of each section's budget to allocate, which compaction tier to select, and which cache policy to apply. Bind performs all I/O, concurrently where possible: memory retrieval, history loading, tool schema resolution, skill loading, and consumption of emergent context carried over from the previous turn. Optimize transforms bound content into a cache-aligned, budget-fitted, provider-specific arrangement, following the deterministic cache-aware policy of Section 3 and constrained by an explicit budget of allowed transformations (reorder, clear, prune, spill, drop). Execute serializes and calls the provider. Feedback records actual input, output, and cache usage in the statistics tables so that future Plan phases are better informed.

\subsection{Phase 1: Plan}

Plan is a pure function of the catalog snapshot. Its central quantity is context pressure,

\begin{equation}
P_{\mathrm{raw}}=\frac{T_{\mathrm{used}}}{L_{\mathrm{eff}}},\qquad
P_{\mathrm{pred}}=\frac{T_{\mathrm{used}}+R_o+R_t+R_s}{L_{\mathrm{eff}}},
\end{equation}

where $L_{\mathrm{eff}}$ is the model's effective input limit and $R_o$, $R_t$, $R_s$ are reserves for the expected model response, extended thinking, and tool-schema growth. The response reserve $R_o$ is not an average but a per-bucket percentile (p75 in steady state, p95 after a recovery event) read from \PipelineStats{} buckets keyed by (model, query source); $R_t$ and $R_s$ are typed reserve slots that the current estimator leaves at zero (Section 3). This is the direct analogue of histogram-based cardinality estimation [22]: averages are too optimistic for long-tail coding turns and invite \code{prompt\_too\_long} recovery loops, whereas a percentile in a typed bucket is robust to workload mix.

$P_{\mathrm{pred}}$ drives a compaction tier from Normal to TrimSchemas, CompactHistory, and AggressivePrune. A gating rule prevents over-compaction: predictive pressure may escalate the tier that raw pressure would select, but never de-escalate. A non-empty \RecoveryState{} (e.g., a recent \code{prompt\_too\_long} error), however, escalates further.

Plan also produces a section manifest: the list of sections, their cache scopes (Global / Session / None), their individual token budgets, and the cache strategy. Every input is a reference into the catalog, which makes Plan deterministic and easy to test independently of the rest of the pipeline.

\subsection{Phase 2: Bind}

Bind fetches the concrete content for each planned section. Because the planned sections are independent, bindings can run concurrently: memory retrieval against an external service, history loading from in-memory state, tool schema resolution, skill loading, and consumption of emergent items from the previous turn all share no write state. The slow bindings are the few that touch network or disk (memory, cloud snapshots, spilled tool results); the remaining bindings are local and inexpensive.

Bind's output is a list of typed artifacts: system-section artifacts, message artifacts, tool-schema artifacts, attachment artifacts, memory artifacts, and spill references. Typing matters because later stages must preserve provider invariants such as tool-call/tool-result pairing, thinking blocks, and media placeholders. String-first assembly hides exactly these invariants. Bind also respects per-section token budgets assigned by Plan.

\subsection{Phase 3: Optimize}

Optimize consists of cache alignment, compaction, and spill decisions. It operates on bound artifacts and produces a fully serializable request and a trace. When the compaction tier selected by Plan calls for it, Optimize has several ways to reduce what is sent: dropping stale tool output, pruning unused schemas, summarizing older history with another model call, or spilling large artifacts behind stable references. These actions differ in cost, reversibility, and risk---dropping a tool result is inexpensive but lossy, summarization preserves intent but may erase details, and spilling keeps bytes out of the prompt while adding retrieval obligations---so a robust optimizer needs not only a current token count but a policy for choosing the least damaging tier, which Section 3 formalizes.

It is constrained by an \OptimizeLimits{} struct that gates each allowed transformation (reorder, clear, prune, spill, LLM summary, round dropping), caps the number of reorder moves per turn, and sets a circuit-breaker on total tokens cleared. Whenever a gate blocks a transformation, the trace records the missed opportunity with its reason, making the optimizer auditable.

Optimize follows the deterministic, cache-aware policy formalized in Section 3. Cache marker placement is delegated to a \ProviderCachePolicy{} declared per backend (prefix-only, Anthropic-style control, Bedrock, OpenAI-compatible), which determines maximum marker count, marker granularity (system block, message, content block), whether global scope is supported, and whether cached tool results can be referenced rather than re-sent. This makes ``how to mark for cache'' a policy decision rather than logic scattered across providers.

\subsection{Phases 4--5: Execute and Feedback}

Execute serializes the optimized artifacts through the provider adapter, optionally stops at EXPLAIN-only mode (returning the trace without calling the model), and otherwise sends the request. Separating Execute from Optimize preserves the property that you can inspect what would have been sent without side effects, in the same way EXPLAIN (without ANALYZE) does for SQL.

Feedback records the actual result: input tokens, cache read, cache creation, output tokens, whether truncation occurred, and an attribution for any cache break. These statistics are kept in \PipelineStats{} buckets that the next Plan phase will read. \PipelineStats{} is the only mutable object during a turn, while the catalog itself is not mutated by the pipeline.

\section{Context Optimization}

This section states the placement problem that the optimizer solves, the deterministic heuristic \ContextPipe{} actually runs, and the invariants that make the output safe for serialization and replay. We deliberately describe the mechanism here rather than an idealized objective, as the optimizer is not a utility-maximizing search but a sequence of gated, order-preserving transforms where every decision, applied or skipped, is recorded in the trace.

\subsection{Problem Statement}

Informally, the optimizer decides which sections to keep, in what order, and where to place cache-marker boundaries, under a hard token budget and a set of sections that should never be dropped.

\heading{Inputs.} Let $S=\{s_1,\ldots,s_n\}$ be the sections produced by Bind. Each section $s_i$ carries an estimated token cost $t_i$, a cache scope $c_i\in\{\mathrm{Global},\mathrm{Session},\mathrm{None}\}$ ordered by volatility (Global is most stable), and a compression priority $p_i$. The priority is either First, Normal, LastResort, or Never, where First sections (the emergent tiers) are compressed before others, while Never sections (Identity, Constraints) are never compressed.

\heading{Decision variables and constraints.} The optimizer makes three decisions: a section order $\pi$, a compaction mask $\mu\in\{0,1\}^n$ (which sections survive), and a set of cache-marker positions $M$, with $|M|\le M_{\max}$ set by the provider policy. Writing $B$ for the effective input budget (model window minus reserves), the decisions must satisfy four constraints:

\begin{itemize}
\item \textit{budget:} retained sections fit, $\sum_{i:\mu_i=1} t_i\le B$;
\item \textit{precedence:} fixed anchors (Identity, Constraints) keep their positions;
\item \textit{protection:} $\mu_i=1$ whenever $p_i=\mathrm{Never}$;
\item \textit{pairing:} every tool-result stays adjacent to its tool-call, as providers require.
\end{itemize}

Maximizing retained utility under a budget with precedence is a precedence-constrained knapsack and is NP-hard in general; adding order-sensitive cache effects further complicates the objective. \ContextPipe{} therefore does not search this space. Instead it applies a fixed, deterministic policy that exploits the structural fact that makes the problem tractable in practice: cost (cache invalidation) and the natural ordering of sections are both governed by the same volatility signal. We choose a stable, auditable cache layout policy that places sections in ascending volatility order.

\subsection{Predictive Context Pressure}

The quantity that drives every compaction decision is context pressure. Given current input tokens $T$ and effective limit $L$,

\begin{equation}
P_{\mathrm{raw}}=\frac{T}{L},\qquad P_{\mathrm{pred}}=\frac{T+R}{L},
\end{equation}

where $R=R_o+R_t+R_s$ aggregates reserves for the expected model response, extended thinking, and tool-schema growth. By construction $P_{\mathrm{pred}}\ge P_{\mathrm{raw}}$, and a zero or invalid limit saturates pressure to 1.0 rather than dividing by zero. The reserve $R$ is estimated from history (Section 3.8); in the current implementation the response reserve $R_o$ is filled from observed completion lengths while $R_t$ and $R_s$ are typed but default to zero, a conservative choice we revisit in Section 9.

Predictive pressure is analogous to histogram-based cardinality estimation in a query optimizer [22]: it lets Plan provision for the turn's likely output before the input is even sent, so the request is compacted before it would have triggered a \code{prompt\_too\_long} error rather than after.

\subsection{Tier-Gated Compaction}

Predictive pressure selects a discrete compaction tier by three fixed thresholds:

\[
\begin{array}{ccl}
0.00\le P<0.60 &:& \text{Normal},\\
0.60\le P<0.75 &:& \text{TrimSchemas},\\
0.75\le P<0.90 &:& \text{CompactHistory},\\
0.90\le P      &:& \text{AggressivePrune}.
\end{array}
\]

The tier is selected from both raw and predictive pressure and the more aggressive of the two is taken, so prediction may only escalate the tier, never relax it (\code{select\_tier\_gated}). A non-empty \RecoveryState{} (e.g., a recent \code{prompt\_too\_long}) escalates the tier further still. This monotone, escalate-only rule prevents the oscillation an averaging estimator would produce on long-tailed coding turns.

Each tier unlocks a specific transform, and each transform is additionally guarded by a boolean in \OptimizeLimits:

\begin{itemize}
\item Normal: no compaction.
\item TrimSchemas: prune tool schemas that the turn is unlikely to use.
\item CompactHistory: clear the oldest tool results, replacing each with a provider-shaped placeholder, under a \code{max\_clear\_tokens} circuit breaker that caps how much can be cleared in one turn; spill oversized results to disk (Section 3.5).
\item AggressivePrune: in addition, drop the oldest whole conversation rounds, with the drop fraction scaling with pressure and a floor of at least four droppable rounds before any drop occurs. System messages are never dropped.
\end{itemize}

When a tier would permit a transform but its \OptimizeLimits{} gate is closed, the optimizer emits a SkippedOptimization trace record naming the step and the reason. The closed-gate case is thus as observable as the applied case, which is the property the audit guarantee below depends on.

\begin{figure*}[t]
\centering
\fbox{\begin{minipage}{0.95\textwidth}\small
\textbf{Algorithm 1: ContextPipe Optimize (deterministic, gated)}\\
\textbf{Input:} bound sections $S$, plan $P$ (tier, budget), provider policy $\Pi$, limits $\mathcal{L}$, latches $\Lambda$\\
\textbf{Output:} optimized request $R$, trace $\tau$
\begin{enumerate}
\item \textbf{Order.} If $\mathcal{L}.\mathrm{reorder}$: sort reorderable sections by $(\mathrm{volatility}(c_i),\mathrm{id}(s_i))$, anchoring Identity and Constraints; adopt only if moves $\le\mathcal{L}.\mathrm{max\_moves}$, else skip and log.
\item \textbf{Compact.} By $P.\mathrm{tier}$ and the matching gate in $\mathcal{L}$: prune schemas (TrimSchemas); clear oldest tool results under the \code{max\_clear\_tokens} breaker (CompactHistory); drop oldest rounds (AggressivePrune). Log every applied and skipped transform.
\item \textbf{Spill.} For each section with $t_i\ge10{,}000$ and a backend present (never Identity/Constraints/WorkingMemory): persist and insert a SpillReference.
\item \textbf{Mark.} Place $\le\Pi.M_{\max}$ markers at scope boundaries, honoring $\Pi$ (prefix-only $\Rightarrow$ none) and $\Lambda$ (suppress None markers on a flipped latch).
\item \textbf{Serialize.} Serialize $(\pi,\mu,M)$ through $\Pi$; append pressure, tier, budget, markers, and all skipped steps to $\tau$. Return $(R,\tau)$.
\end{enumerate}
\end{minipage}}
\end{figure*}

\subsection{Cache-Aware Ordered Placement}

The reordering step (\code{cache\_align\_sections}) is a pure sort, not a search. Reorderable sections are sorted by ascending cache-scope volatility, Global $<$ Session $<$ None, with a stable section identifier breaking ties so the result is a total, deterministic function of $S$. The fixed anchors Identity and Constraints are excluded from the sort and keep their leading positions. To bound churn, the new order is adopted only if it differs from the current order by at most \code{max\_reorder\_moves} transpositions; otherwise the sort is skipped and logged.

Ascending volatility is central to the cache design. Cache markers are then placed at scope boundaries, the point where a run of Global/Session sections gives way to None. Because every byte before the deepest marker is stable across the session, the marked prefix is byte-identical turn over turn and is served from the provider's prefix cache [2, 14, 29]. A single volatile section placed early would push the boundary backward and shrink the cacheable prefix; the sort guarantees this never happens. Marker placement respects three further constraints: a prefix-only provider policy emits no markers at all; if a \SessionLatches{} value flipped this turn, None-scope markers are suppressed to avoid caching content that changed in the current turn; and the total marker count is capped at $M_{\max}$.

\subsection{Spill}

When a single section exceeds a fixed threshold (10,000 tokens) and a spill backend is configured, its content is persisted to the backend and replaced in-place by a lightweight SpillReference that frees the token budget while preserving a recovery path. Identity, Constraints, and Working Memory are never spilled. Rehydration is fail-open: if the backend cannot return spilled content, the optimizer substitutes a placeholder and records a skipped-rehydration trace entry rather than aborting the turn, keeping the reference for a later retry.

\subsection{The Heuristic End-to-End}

Algorithm 1 describes how the Optimizer works, where the sort costs $O(n\log n)$ and compaction, marking, and serialization $O(n)$; with $n\le30$ sections and a few dozen messages the optimizer is negligible beside one LLM call. The optimization should be a total, deterministic, fully-audited function of its inputs, resulting in three properties formalized below.

\subsection{Invariants and Guarantees}

The optimizer is safe because it preserves a small set of invariants that the serializer and providers rely on. We state them over the output of Algorithm 1.

\heading{Lemma 1 (Anchor and precedence).} Identity and Constraints retain their leading positions, and no reorder crosses a cache-scope class boundary.

\textit{Proof.} Step 1 excludes the anchors from the sort and sorts the remainder by a key whose primary component is the scope class; no other step changes $\pi$. $\square$

\heading{Lemma 2 (Protection).} For every $i$ with $p_i=\mathrm{Never}$, $\mu_i=1$.

\textit{Proof.} The mask starts at 1; each compaction transform in step 2 guards against clearing Never sections, and step 3 excludes Identity/Constraints/WorkingMemory from spill. $\square$

\heading{Lemma 3 (Pairing and recoverability).} Every serialized tool-result is preceded by its tool-call, and every cleared oversized result leaves a recoverable reference.

\textit{Proof.} Compaction carries tool-call/result pairs as atoms; the serializer rejects an unpaired result. Spill writes the backend before replacing content, and a backend failure downgrades to a logged placeholder rather than data loss. $\square$

\heading{Property 1 (Audit completeness).} For every turn, $\tau$ records each applied transform and each skipped transform with its reason, so the final $(\pi,\mu,M)$ is reconstructable from $\tau$ and the inputs.

\textit{Proof.} Steps 1--4 write to $\tau$ on both the applied and the gated-skip branch; the only unlogged steps (the initial sort key, the identity mask) are deterministic functions of the inputs. $\square$

\heading{Property 2 (Replay determinism).} Given the same \ContextSources{} snapshot, the same \PipelineStats{} snapshot, the same \OptimizeLimits{} and provider policy, Algorithm 1 produces a byte-identical $R$ and $\tau$.

\textit{Proof.} The algorithm has no randomness: the sort key is a total order, tier selection is a pure function of pressure and limits, and marker placement follows a deterministic policy. Statistics are read-only during the turn. Feedback writes only after Execute (Section 3.8), so the snapshot read at Plan time is immutable throughout. Serialization is a function of $(\pi,\mu,M,\Pi)$. $\square$

\heading{Property 3 (Failure isolation).} If Execute of turn $N$ fails, the persistent state, consisting of \PipelineStats, the spill store, and the \EmergentContext{} queue for turn $N+1$, is unchanged except for an explicit failure record.

\textit{Proof.} Feedback is the only writer of usage statistics and runs only after a successful Execute; on failure it writes a single diagnostic record and no usage sample, so percentile buckets are not poisoned by a non-response. Spill writes are atomic, and \EmergentContext{} for $N+1$ is populated only on success. $\square$

We state these as properties rather than theorems: each follows directly from the pipeline construction, and their value is as tested specifications of the implementation, not as mathematical results. Property 1 makes the pipeline auditable; Property 2 makes shadow-rollout diffing and post-mortem replay well-defined; Property 3 makes retry safe without bespoke cleanup. Crucially, determinism here is stronger than it would be under a utility-search optimizer, precisely because the policy is a fixed sort rather than a data-dependent search.

\subsection{The Statistics Subsystem}

\PipelineStats{} plays the role of the catalog statistics store in a query optimizer: read-only to Plan, write-only to Feedback, with no read--write interleaving inside a turn.

\heading{Reserve estimation.} Response reserves are estimated per bucket keyed by (model, \code{query\_source}), so a chat workload does not distort the reserves a coding agent uses. Each bucket holds a PercentileDigest: a sorted sample list capped at 512 entries that, once full, evicts the median on each insert so that the tail quantiles driving reserves stay accurate. Plan reads p75 in steady state and p95 while in recovery; an empty bucket returns a fixed floor (500 tokens). We use an exact sorted digest rather than an approximate streaming sketch because per-turn sample volume is small and the implementation is simpler to audit; the digest exposes an \code{is\_saturated} flag so that a future migration to a streaming quantile estimator can be triggered by observed data.

\heading{Cache and section tracking.} A single exponentially weighted moving average of the cache-hit ratio (seeded on turn one, then $\lambda=0.1$) summarizes how well the prefix cache is performing. Per-section token usage is tracked by a separate EMA ($\alpha=0.3$) and feeds the next turn's budget allocation. Each section's rendered content is fingerprinted, and a per-kind churn counter increments whenever a kind's content changes; this is how the runtime learns which nominally stable sections are unexpectedly volatile and therefore poor cache anchors. Cache breaks and compaction events are kept in 64-entry ring buffers; a compaction cascade (two events in three turns, or three in ten) is surfaced as an alert so operators can see a context that is growing faster than compaction can stabilize it.

\heading{Failed-turn and concurrency semantics.} Following Property 3, a failed Execute contributes a failure record but never a usage sample, so an aborted turn cannot poison the percentile buckets. Bind performs all I/O concurrently but reads only an immutable snapshot; Feedback mutates statistics after Execute returns, outside the concurrent fan-out. The pipeline therefore exhibits turn-level serializability: Plan of turn observes exactly the writes of Feedback of the previous turn and nothing else.

\section{ContextSources: The Pipeline Catalog}

This section describes \ContextSources, the catalog abstraction used by the pipeline to reason about available context (Table~\ref{tab:lifecycle}).

\begin{table}[t]
\caption{Context data sources by lifecycle. The two emphasized tiers are novel in \ContextPipe; production systems typically handle them ad hoc.}
\label{tab:lifecycle}
\centering\scriptsize
\begin{tabularx}{\columnwidth}{@{}lXl@{}}
\toprule
Tier & Examples & Bind cost\\
\midrule
Immutable & core rules, output format & free\\
Per-agent & tool schemas, skill catalog & free\\
\textbf{Latched} & beta headers, cache scope & free\\
Per-session & edge profile, project rules, self-model & free/I/O\\
Per-turn & messages, facts, active skills & free\\
External & memory, spill, cloud snapshots & remote I/O\\
\textbf{Emergent} & prefetched memory, skill discovery, tool-use summaries, attachments & free\\
Feedback & cache stats, token histograms & free\\
\bottomrule
\end{tabularx}
\end{table}

\begin{table}[t]
\caption{Catalog record schema. Every data source registers a record with these fields; Plan reads the schema to decide what to bind.}
\label{tab:schema}
\centering\scriptsize
\begin{tabularx}{\columnwidth}{@{}lX@{}}
\toprule
Field & Meaning\\
\midrule
\code{tier} & one of eight lifecycle tiers (Table~\ref{tab:lifecycle})\\
\code{kind} & section kind (e.g., Identity, Memory, History)\\
\code{scope} & cache scope: Global / Session / None\\
\code{priority} & compression priority: Never, LastResort, Normal, First\\
\code{est\_tokens} & predicted bind cost (per-kind usage EMA)\\
\code{bind\_latency} & measured I/O time, recorded per bound section\\
\code{recoverable} & whether the source could be re-fetched if lost\\
\code{provider\_markerable} & whether the section could host a cache marker\\
\code{dependencies} & other catalog kinds this source reads\\
\bottomrule
\end{tabularx}
\end{table}

A query optimizer is only as good as its catalog. In most agentic runtimes, the implicit catalog is a flat struct that has accumulated roughly 200 fields mixing per-turn ephemera, per-session state, per-agent configuration, and cross-session learning. Plan cannot reason about what to fetch because there is no enumeration of sources.

\ContextSources{} classifies every data source by lifecycle (how often it changes), location (where it lives), and bind cost (how expensive it is to fetch). Plan queries the stable tiers to decide what to fetch, while Bind incurs the I/O cost only for the tiers Plan selects. Each catalog entry carries a fixed schema (Table~\ref{tab:schema}), so Plan could enumerate what is available without reflection. This schema is the direct analogue of \code{information\_schema.columns} in a relational catalog that makes the data queryable.

Two tiers deserve emphasis because they are the main source of hard-to-trace behaviors in existing runtimes.

\heading{SessionLatches.} A latch is a value that is evaluated lazily on the first trigger and then frozen for the session. Beta headers, cache-scope eligibility, and provider feature flags are the usual examples. Latches exist precisely to prevent cache breaks: if a header could toggle mid-session, it would invalidate the KV cache prefix, and rolling that back would be expensive. This is the context analogue of a one-shot DDL committed inside a transaction. By modeling latches as a dedicated tier, the Optimizer could reason about whether a latch has already fired, whether the prefix hash has therefore changed, and whether a cache break this turn is attributable to the latch.

\heading{EmergentContext.} Although the pipeline is staged, context discovery could continue after Plan. Tool execution produces new context that was not predictable at Plan time: writing file $X$ may trigger discovery of skill $Y$; a small-model auxiliary call may generate a tool-use summary; memory prefetches may complete during streaming. These items are queued at the end of the current turn and consumed at the start of the next turn. This is the context analogue of adaptive query execution [3, 28], where the executor discovers mid-flight that row estimates were wrong.

Because emergent items are produced after the current Plan but consumed by the next Bind, they carry a safety triple: a per-turn TTL (default: consumed on the immediately next turn only), a content hash for deduplication at write time, and a per-list cap. Without these guardrails, resume/replay could silently double-inject stale attachments.

\subsection{Runtime Introspection}

At runtime, the system needs to return catalog schemas (Table~\ref{tab:schema}). Together these form the context analogue of EXPLAIN, and EXPLAIN (COSTS ON) in a relational system. An operator debugging a regressed session could ask ``what would the pipeline do right now?'' without waiting for the next user turn. This gives an operator a procedure for introspecting the pipeline without redeploying or attaching a debugger.

\subsection{How Stages Use the Optimizer State}

Table~\ref{tab:state} binds the pipeline stages (Section 2) to the optimizer state of Section 3. Each stage reads or writes a specific, disjoint subset, which makes the pipeline unit-testable: a synthetic snapshot of reserves and a section list is enough to drive any stage in isolation, without a live provider.

\begin{table}[t]
\caption{Pipeline stages $\leftrightarrow$ optimizer state.}
\label{tab:state}
\centering\scriptsize
\begin{tabularx}{\columnwidth}{@{}lXl@{}}
\toprule
Stage & State read/written & Invariants\\
\midrule
Plan & reserves $R_o$, pressure $P_{\mathrm{pred}}$, tier, $M_{\max}$ & ---\\
Bind & per-section actual $t_i$, skips by budget & ---\\
Optimize & order $\pi$, mask $\mu$, markers $M$, limits $\mathcal{L}$ & L1--L3, P1--P2\\
Execute & serializes $(\pi,\mu,M)$ via canonical serializer & L3\\
Feedback & cache-hit EMA, percentile samples, churn, breaks & P3\\
\bottomrule
\end{tabularx}
\end{table}

\section{Explain Analyze}

This section describes the EXPLAIN ANALYZE trace emitted by \ContextPipe{} and how it supports auditability, replay, and feedback.

\begin{figure}[t]
\begin{lstlisting}
-- EXPLAIN ---------------------
ContextPlan
 pressure: 0.72 (raw: 0.65)
 reserves: out=480 think=0 sch=0
 tier: CompactHistory
 cache: AnthropicCacheControl (2)

Sections (plan to actual tok)
 (1) Identity [G] 480 (keep)
 (2) SelfModel [S] 320 (last)
 (3) ProjectContext [S] 150 (last)
 (4) Memory [-] 890/1200 (4)
 (5) History [-] 3200/5000 (12)
 (6) Constraints [G] 280 (keep)
 Skills:code_review [S] 650
 RuntimeIdentity [-] 120

Compaction
 tier=CompactHistory, keep=2, budget=4000
 cleared 5 results (3200 to 850), 3 spilled

Cache markers
 m[0]: after (1)+(6)       G ( 760)
 m[1]: after (2)+(3)+Sk   S (1880)

Total: 6090 / 8500 eff (71.6% util)

-- ANALYZE (post-execution) ----
Actual input:    6234 (err +2.4%)
Cache read:      1880 (m[0]+m[1] hit)
Cache creation:     0 (full prefix hit)
Cache hit ratio: 100% (avg: 87%)
Output tokens:    520 (est 480, +8.3%)
\end{lstlisting}
\caption{EXPLAIN ANALYZE exposes both planning decisions and post-execution cache behavior for one turn. EXPLAIN is the pre-execution trace; ANALYZE is the post-execution feedback joined back in. Region tags: G=Global, S=Session, -=None.}
\label{fig:explain}
\end{figure}

Every turn produces a structured trace (Figure~\ref{fig:explain}). The pre-execution part mirrors EXPLAIN: it shows the pressure decomposition, the selected tier, per-section planned and actual tokens, compaction decisions, and cache marker positions. The post-execution part mirrors ANALYZE: it records actual input and output tokens, cache read and creation tokens, and deltas against the plan's estimates.

This trace is not merely an auxiliary debugging artifact, but also the pipeline's audit log: every optimizer decision, every skipped transformation, every cache marker, every emergent injection, and every recovery escalation is visible on the turn it happens. Eight alert rules run against each trace: explicit prompt-cache break (with attribution), cache cold-start on turn $>1$, cache-hit regression over a moving window, predictive miss above 20\%, compaction cascade, recovery loop, pressure spike in one turn, and emergent-list overflow. A plan/execute mismatch is additionally reported as a validation event at the stage boundary rather than as an alert rule. Warnings surface in the REPL/admin UI; errors could escalate into the next Plan (e.g., a recovery loop forces AggressivePrune). Thus unexpected behavior is recorded in the trace rather than appearing as silent degradation. Figure~\ref{fig:explain} illustrates an unedited excerpt; a real exported trace appears in Table~\ref{tab:trace}, where each compaction-tier crossing is visible as a one-call cache-break dip.

\section{Implementation}

This section describes how \ContextPipe{} is grounded in an implementation while preserving the properties in Section 3.7.

\heading{Components.} There are two main components: pipeline data types, and runtime orchestration that reads live agent state, performs I/O, and dispatches tool calls. The pipeline data types include pressure, reserves, catalog snapshots, provider cache policy, trace records, and typed artifacts. The runtime takes the pipeline inputs and feeds ContextFeedback and \EmergentContext{} back into the next pipeline invocation.

\heading{Typed artifacts.} Internally, every section is a typed artifact rather than a String. A SystemSectionArtifact carries its kind, cache scope, priority, and content; a MessageArtifact preserves tool-call identifiers and role metadata; a ToolSchemaArtifact carries the schema plus a stable hash for cache-break attribution; a SpillReference carries the tool-call id, byte size, and disk path. Serialization to the provider's wire format happens exactly once, at the Execute boundary. This single-serializer ownership discharges Lemma 3 (pairing and recoverability) and prevents an optimizer pass from accidentally dropping thinking blocks, orphaning tool results, or reshaping media placeholders.

\heading{Stage-boundary validation.} The implementation treats every stage boundary as a contract, not just as a function call. Execute receives both the optimized request and the trace record emitted by Optimizer, then checks that the serialized request still matches the selected tier, cache-marker count, compaction actions, and protected-section set. A mismatch is recorded as a first-class trace event and blocks provider dispatch in shadow mode. This catches the failure class where a planner or the Optimizer makes a decision that is silently ignored downstream, collapsing a staged pipeline back into a flat assembler while leaving aggregate metrics ambiguous.

\heading{Failure and recovery matrix.} Every known failure mode maps to a pipeline response. A \code{prompt\_too\_long} error escalates \RecoveryState{} and forces a harder tier on the next Plan. A \code{max\_output\_tokens} retry bumps the output reserve for that (model, query-source) bucket. A sudden cold cache causes Feedback to diff the prompt/tool/model/latch hashes against the previous turn and record the attribution before any reordering is attempted. A summarizer overflow falls back to drop the oldest compactable rounds and then truncation, each step recorded in the trace. Memory retrieval failure surfaces as missing memory in the trace rather than as an empty success. Each of these is one table row in the pipeline's failure matrix, not scattered across the runtime.

\subsection{ForkPrefix: Cache-Prefix Sharing for Parent-Child Agents}

Multi-agent delegation exposes the cache-alignment problem at a higher level: a parent agent spawns a child, and both agents should share the cache prefix they have in common (system rules, tool schemas, project context). Without sharing, the child's first turn is a full cache miss.

ForkPrefix solves this by capturing a frozen, byte-identical snapshot of the parent's cacheable request prefix at the moment of spawn. The snapshot carries (i) the canonical wire bytes of the prefix region (the frozen system blocks plus tool schemas), produced by the runtime's own serializer rather than by an SDK so that SDK version drift cannot perturb the hash; (ii) a per-tool canonical schema with its own SHA-256, so a later cache break could be attributed to exactly which tool's description churned; and (iii) the elements that the provider folds into its cache key: the thinking-config slice, the provider/model identity, and the sorted beta headers. A single SHA-256 over the canonical prefix bytes is the authoritative record of whether anything drifted between capture and use. A first-turn cache probe compares the child's observed \code{cache\_read\_tokens} against the parent-side estimate carried on the resolved snapshot and emits one audit event per spawn, so a cache-sharing failure is visible even before an executor fully consumes the snapshot. Full child-side consumption through the same pipeline's Bind phase---the step that turns an inherited snapshot into the child's actual first-turn prefix, rather than only an audited estimate---is not yet wired in our default server executor, which runs delegated children fresh today; by design, an executor that skips this field degrades to pre-ForkPrefix behavior rather than failing.

Three properties of the mechanism as designed echo structures from query optimization; each holds once an executor consumes the snapshot. First, the snapshot is immutable after capture: the system blocks are read-only with no public mutation API, so content the parent adds after spawn (for example, a microcompaction) does not retroactively change the child's prefix, and a skill that needs extra guidance must append it as a user-message suffix that sits after the cached prefix. Second, a \code{validate\_spawn} step is a precondition for cache sharing: it rejects a spawn whose provider, model, thinking budget, or prefix size would break the captured cache identity (e.g., a child \code{max\_output\_tokens} that clamps the thinking budget to a different value), returning a typed error so the caller could fall back rather than silently miss. Third, the snapshot supports a skip-cache-write mode for side-queries: forked children that are auxiliary (e.g., a small-model summarization call) reuse the prefix without writing a fresh cache entry at their own tail, by shifting the final marker one position upstream. This is the context analogue of a shared plan cache across concurrent queries, and it generalizes naturally to peer-to-peer sibling sharing as future work.

\heading{Shadow-pipeline rollout.} Any change to context assembly risks breaking KV-cache alignment, so changes cannot be rolled out with a single feature-flag switch. \ContextPipe{} ships with a mandatory shadow mode: the new pipeline runs alongside the active assembly path on every turn, but only the active path sends a request. A diff checker compares system-block byte hashes, tool schema hashes, message count and role sequence, cache marker positions, and total token estimates. Rollout proceeds in four steps: shadow-only, shadow-verified (zero diff for $N$ turns), flip (new path serves, old becomes shadow, catching regressions), and retire (old path removed after $M$ sessions with zero diff). The shadow pipeline is how an optimizer becomes safe to iterate on.

\section{Evaluation}

This section evaluates \ContextPipe{} using the backbone model DeepSeek-V4-Pro [26].

We compare two context construction policies: \textit{Structured} is the full \ContextPipe{} pipeline, while \textit{Flat} forces every optimizer gate closed (\code{all\_closed}): sections keep binding order that keeps a stable prefix, without pruning, compaction, or spills. Tier selection is still computed in both, due to the shared planner, so any tier that Flat reports is a label, not an action, which gives a per-call record of when the mechanism would have engaged.

\heading{Workload.} We choose the Qutebrowser subset of SWE-bench Pro [8], the same subset used by Tressoir [18]. SWE-bench Pro is a contamination-resistant successor to SWE-bench [13], and the subset is long-horizon and touches multiple files, where each instance pairs a bug report with the codebase state before the fix and a test suite that validates a correct patch. We preliminarily evaluated 3 of 79 instances in the subset to stress the context window, the prompt cache, and the compaction ladder at once, and plan to run more instances as future work. Each instance runs three times.

\begin{table}[t]
\caption{Context optimizer on SWE-bench Pro Qutebrowser (DeepSeek-V4-Pro; 3 instances $\times$ 3 repeats per condition, interleaved matched pairs; per-cell means). Changes is $(\mathrm{Flat}-\mathrm{Structured})/\mathrm{Flat}$.}
\label{tab:evaluation}
\centering\scriptsize
\begin{tabular}{@{}lrrr@{}}
\toprule
 & Flat & Structured & Changes\\
\midrule
Total prompt tokens & 1,045,222 & 730,978 & 30.1\%\\
Fresh (uncached) tokens & 46,581 & 99,967 & $-114.6$\%\\
Cache-hit rate & 95.3\% & 86.2\% & $-9.1$\%\\
Duration (s) & 204 & 188 & 7.8\%\\
LLM calls & 30.2 & 25.0 & 17.2\%\\
Optimizer actions & 0 & 2,417 & ---\\
\bottomrule
\end{tabular}
\end{table}

\heading{Structured acts, at the tradeoff of lower cache-hit ratio.} Table~\ref{tab:evaluation} demonstrates token usage and other metrics between Structured and Flat. Comparing to Flat, Structured reduces 30\% total tokens, 39\% repeated cache reads, 23.5\% completion tokens, 23.1\% LLM calls, and 8.7\% response time. These reductions come at a cost in lower cache-hit ratio: fresh (uncached) input token doubles, and the per-cell cache-hit median drops from 95.6\% to 86.3\%, because every compaction changes the sent prefix. Decomposing billed cost as

\begin{equation}
E(r)=\mathrm{fresh}+r\cdot\mathrm{cached},
\end{equation}

where $r$ is the provider's cached-token price ratio, the break-even is $r^*=0.145$: at a DeepSeek-like $r\approx0.1$ the Flat baseline incurs approximately 11\% lower billed cost despite sending 43\% more total context, while Structured incurs the lower billed cost wherever cached tokens cost more than 14.5\% of fresh (e.g., $-13$\% at $r=0.25$). Structured's advantage in this regime therefore lies in context volume, calls, response time, and window headroom rather than unconditionally in billed cost.

\begin{table}[t]
\caption{Real EXPLAIN ANALYZE trace excerpt (Structured, highest-pressure instance, repeat 1; per-call cache-hit from post-response feedback).}
\label{tab:trace}
\centering\scriptsize
\begin{tabular}{@{}rrrrl@{}}
\toprule
Call & $P_{\mathrm{raw}}$ & $P_{\mathrm{pred}}$ & Cache hit & Tier\\
\midrule
1 & 0.05 & 0.06 & 24\% & Normal\\
2 & 0.06 & 0.06 & 98\% & Normal\\
$\cdots$ & & & &\\
12 & 0.20 & 0.20 & 99\% & Normal\\
13 & 0.20 & 0.20 & 8\% & TrimSchemas\\
14 & 0.20 & 0.20 & 100\% & TrimSchemas\\
$\cdots$ & & & &\\
23 & 0.29 & 0.29 & 96\% & TrimSchemas\\
24 & 0.31 & 0.32 & 11\% & CompactHistory\\
25 & 0.32 & 0.32 & 99\% & CompactHistory\\
\bottomrule
\end{tabular}
\end{table}

\heading{A debugging case study, from the same trace.} Table~\ref{tab:trace} is an unedited excerpt of the exported per-session trace of one Structured cell (the highest-pressure instance, repeat 1), and it illustrates the operator-facing diagnosis the trace is designed to support. The cold start is visible (call 1: 24\% hit). The TrimSchemas crossing at call 13 appears as a one-call cache-break (8\% hit, because schema pruning changed the prefix bytes), followed by immediate recovery to 93--100\% for the next ten calls. The CompactHistory escalation at call 24 shows the same signature (11\%), recovering to 99\% on the very next call, while $P_{\mathrm{raw}}$ stays bounded at 0.33 through the end of the session. Every dip is attributable to a specific, logged optimizer decision; in Flat's trace of the same instance the hit column stays at 77--100\% after the cold start and the pressure column climbs monotonically until the run hits the max-turn limit.

\heading{Metrics.} We report quality as resolve rate with evaluator validity gates (patch applies, tests complete, no test files modified); cost as prompt/completion tokens, provider-normalized fresh/miss and cache-read tokens, API-call counts, and wall-clock duration; and mechanism as per-call tier decisions, schema prunes, message compactions, spills, and per-call cache-hit from post-response feedback, all exported from the session trace (Section 5). Cache fields are provider-normalized because protocols differ; cross-provider counters are comparable diagnostics, not accounting identities.

\heading{Ablation status.} The Structured-vs-Flat comparison above is the first of the mechanism ablations we plan. Still open, each holding the rest of \ContextPipe{} fixed: (A1) predictive vs. reactive pressure (the two signals differ by at most 1.7\% of the window here, so separating them requires a knob where tier decisions sit near a threshold); (A2) \SessionLatches{} vs. per-turn re-evaluation; (A3) TTL+dedup+cap on \EmergentContext{} vs. append-only; and (A4) \ProviderCachePolicy{} vs. fixed marker placement.

\heading{Limitations and threats to validity.} (i) The experiments cover only 3 of 79 instances in the Qutebrowser subset, which limits broad implications of \ContextPipe. (ii) The token cost model per LLM also significantly impacts the cost effectiveness of this work. (iii) Cache counters are provider-normalized. (iv) Mechanism ablations A1--A4 remain open.

\section{Related Work}

Prior work touches pieces of the context-assembly problem without addressing the composition step directly: retrieval and long-context modeling decide what content is available, prompt caching and serving systems determine what a given byte layout costs, and agent loops and orchestration frameworks generate the state that must be assembled each turn. \ContextPipe{} composes selected content onto that priced substrate under a hard window budget, every turn, and borrows its discipline from relational query evaluation, discussed last.

\heading{Retrieval-augmented generation and long context.} Classical RAG [12, 15] and its learned variants (Self-RAG [4], adaptive retrieval, FiD, RAG-Token, Gepa [1]) decide what to retrieve; query-rewriting approaches such as HyDE [9] and Causal Prompt Optimization [6] improve the query itself before dispatch. Long-context models [5] push how much context can be consumed, and recent work explores RL- or RLHF-guided selection of what to include in a long-context window [4, 15]. \ContextPipe{} does not train such a policy; it records the optimizer's raw signals---per-section token usage, per-boundary cache hit/miss, context pressure, and tier decisions---in the per-turn trace, so a learned policy could consume them offline without changing the runtime contract. These techniques plug into Bind's memory sources: they decide candidates, while \ContextPipe{} decides how the candidates fit together under a cache budget.

\heading{Agent memory and compaction.} MemGPT [20] is the most direct antecedent: it formalizes tiered memory between a ``main context'' and an external store, paging on token thresholds. \ContextPipe{} catalogs eight lifecycle tiers rather than one store, drives tier selection from predictive, percentile-based pressure rather than reactive thresholds, and treats cache alignment and provider policy as first-class planning concerns, which MemGPT does not. Production agent runtimes also embed bespoke compaction routines (micro-compact, snippet, reactive); \ContextPipe{} subsumes these under one gated optimizer with an audit trail. PEEK [11] approaches similar pressure from an operating-systems angle, using a context map as an orientation cache for long-context agents; \ContextPipe{} instead frames context management as a database-style execution problem, with a catalog, optimizer, provider cache policy, and per-turn EXPLAIN trace. Both recognize that long-horizon agents need structured state beyond flat transcript concatenation, and differ mainly in the systems abstraction used to organize it.

\heading{Prompt caching and LLM serving.} Automatic prefix caching [19, 29] and explicit cache control [2] provide the underlying mechanism; PagedAttention [14] manages the KV memory, and attention kernels [7] set the raw serving cost. These systems price a byte layout; \ContextPipe{} chooses it---where to place stable content, where to place breakpoints, and how to shape placeholders so byte-prefixes survive compaction---parameterized by a \ProviderCachePolicy{} rather than provider-specific glue.

\heading{Agent loops and benchmarks.} ReAct [27], Reflexion [23], Toolformer [21], LATS [30], and Voyager [24] define what the agent should do; evaluation suites such as SWE-Bench [13], AgentBench [17], and SWE-bench Pro [8] drive the tasks these loops are judged on. Both are orthogonal to \ContextPipe, which runs inside any such loop on every API call; end-to-end agent-success ablations on these benchmarks are future work (Section 9).

\heading{Agent frameworks and orchestration.} Application toolkits such as LangChain, LlamaIndex, and Haystack make it easy to build an agent quickly, not to optimize the per-turn request under a cache and pressure budget; an application built on any of them could adopt \ContextPipe{} as its request-assembly layer without changing its topology. Multi-agent orchestration layers such as AutoGen [25] and CrewAI focus on how agents talk to each other, not on what each agent sees per turn; their abstractions sit above our pipeline, and ForkPrefix-style cache sharing (Section 6) lets participating agents cooperate without $O(\mathrm{agents})$ cache creation. Tressoir [18] similarly studies long-running multi-agent systems through an interpretable blueprint for designing and evolving agent architectures, prompts, tools, and knowledge, but leaves the per-turn context-assembly problem implicit: it exposes no database-style catalog, optimizer, provider cache policy, or EXPLAIN trace. \ContextPipe{} occupies that narrower layer instead of proposing a new multi-agent design language.

\heading{Query optimization.} Classical query optimization [10, 22] and adaptive query execution [3, 28] give our pipeline its vocabulary and shape. The context problem differs in two ways: we cannot precisely quantify ``information value per token,'' so the optimizer is heuristic and feedback-driven rather than cost-based in the Selinger sense, and the consumer (the LLM) is intelligent enough to partially compensate for suboptimal context. \ContextPipe{} respects both differences by making decisions auditable, reversible, and gated, while still borrowing staged planning, statistics, buffer pools, and EXPLAIN.

\section{Conclusion and Future Work}

We presented \ContextPipe, constructing context for long-horizon LLM agents as a query execution problem: a five-phase pipeline binds session states from a lifecycle-indexed catalog, a deterministic cache-aware optimizer places and compacts the bound content under explicit gates, and an EXPLAIN ANALYZE tracker records the result. We showed that this construction makes prompt assembly auditable, replayable, and failure-isolated.

Future work follows directly from the measured total-token vs. fresh-token/cache-hit-ratio tradeoff on more benchmark cases. Finer-grained compaction targets its cause: the tier-gated policy compacts a whole eligible chunk once a threshold is crossed, so each escalation appears in the exported trace as a single large cache break; compacting smaller increments earlier would spread the same savings over smaller breaks, and the predictive-vs-reactive pressure ablation (A1) provides a direct test of whether anticipated pressure enables this. Adaptive cache-marker placement targets its impact: markers are currently placed by a fixed \ProviderCachePolicy{} at scope boundaries, and learning positions from per-boundary hit/miss history could keep a compaction from invalidating stable markers upstream of it, reducing the fraction of the prefix that each break resends.

Beyond this tradeoff, we plan to close the remaining mechanism ablations (A2--A4) using the shadow-pipeline infrastructure; to run larger task panels, drawn from additional benchmarks and repositories beyond SWE-bench Pro Qutebrowser, that turn the exploratory quality tie into a powered non-inferiority comparison and test whether the measured effects hold outside a single codebase; to compare open-source and production coding harnesses under matched budgets with the same model, retry, and caching policies; and, longer term, to generalize ForkPrefix from parent-to-child prefix reuse to peer-to-peer sharing among sibling sub-agents.

\end{document}